# Learning Rigidity in Dynamic Scenes with a Moving Camera for 3D Motion Field Estimation

Zhaoyang Lv[1*], Kihwan Kim[2], Alejandro Troccoli[2], Deqing Sun[2], James M. Rehg[1], Jan Kautz[2]

[1] Georgia Institute of Technology, Atlanta, U.S.
{zhaoyang.lv,rehg}@gatech.edu
[2] NVIDIA, Santa Clara, U.S.
{kihwank,atroccoli,deqings,jkautz}@nvidia.com

**Abstract.** Estimation of 3D motion in a dynamic scene from a temporal pair of images is a core task in many scene understanding problems. In real-world applications, a dynamic scene is commonly captured by a moving camera (i.e., panning, tilting or hand-held), increasing the task complexity because the scene is observed from different viewpoints. The primary challenge is the disambiguation of the camera motion from scene motion, which becomes more difficult as the amount of rigidity observed decreases, even with successful estimation of 2D image correspondences. Compared to other state-of-the-art 3D scene flow estimation methods, in this paper, we propose to *learn* the rigidity of a scene in a supervised manner from an extensive collection of dynamic scene data, and directly infer a rigidity mask from two sequential images with depths. With the learned network, we show how we can effectively estimate camera motion and projected scene flow using computed 2D optical flow and the inferred rigidity mask. For training and testing the rigidity network, we also provide a new semi-synthetic dynamic scene dataset (synthetic foreground objects with a real background) and an evaluation split that accounts for the percentage of observed non-rigid pixels. Through our evaluation, we show the proposed framework outperforms current state-of-the-art scene flow estimation methods in challenging dynamic scenes.

## 1 Introduction

The estimation of 3D motion from images is a fundamental computer vision problem, and key to many applications such as robot manipulation [3], dynamic scene reconstruction [16,25], autonomous driving [9,29,31,46], action recognition [45], and video analysis [15]. This task is commonly referred as *3D motion field* or *scene flow estimation*. 3D motion field estimation in a dynamic environment is, however, a challenging and still open problem when the scene is observed from

---
[*] This work started during an internship that the author did at NVIDIA.

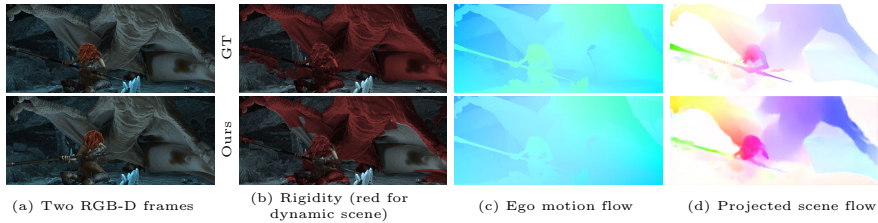

(a) Two RGB-D frames    (b) Rigidity (red for dynamic scene)    (c) Ego motion flow    (d) Projected scene flow

Fig. 1: Our estimated Rigidity (b), Ego-motion Flow (c) and Projected scene flow (d) (bottom row) compared to the ground truth (top row). The rigidity mask allows us to solve for the relative camera transform and compute the 3D motion field given the optical flow.

different view points and the amount of coverage of moving objects in each image is significant. This is mainly because the disambiguation of camera motion (ego-motion) from object motion requires the correct identification of *rigid static structure* of a scene. Unlike other methods solving the problem with piecewise rigid motion [43,21,10], clustering local motions [18], and semantic segmentation [34,47], our network can infer per-pixel rigidity by jointly learning rigidity and the relative camera transform from large-scale dynamic scene data. A brief example of our results is shown in Fig. 1.

Our framework, shown in Fig. 2, takes a sequential image pair with color and depth (RGBD) as the input and mainly focuses on dynamic scenes with a moving camera (e.g., panning), where camera motion and objects motions are entangled in each observation. To solve for 2D correspondences, our framework relies on 2D optical flow, and is not tied to any particular algorithm. We use the method by Sun et al. [35], which we evaluate together with the rigidity network to estimate both ego-motion and scene-motions. The network that learns the per-pixel rigidity also solves for the relative camera pose between two images, and we can accurately refine the pose as a least square problem with the learned dense flow correspondences and rigidity region. To provide better supervision during training and encourage generalization, we develop a tool and methodology that enables the creation of a scalable semi-synthetic RGB-D dynamic scene dataset, which we call *REFRESH*. This dataset combines real-world static rigid background with non-rigid synthetic human motions [38] and provides ground truth color, depth, rigidity, optical flow and camera pose.

In summary, our major contributions are:

1. A learning-based rigidity and pose estimation algorithm for dynamic scenes with a moving camera.
2. An RGBD 3D motion field estimation framework that builds on inference from rigidity, pose, and existing 2D optical flow, which outperforms the state-of-the-art methods.
3. A new semi-synthetic dynamic scene data and its creation tool: REal 3D From REconstruction with Synthetic Humans (REFRESH).

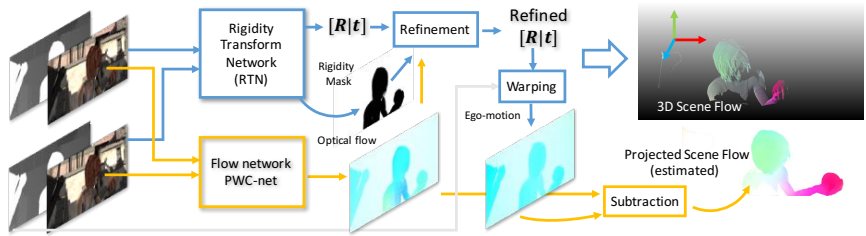

Fig. 2: **An overview of our proposed inference architecture for 3D motion field estimation.** Our method takes two RGB-D frames as inputs independently processed by two networks. The Rigidity Transform Network (RTN) estimates the relative camera transform and rigid/non-rigid regions. The flow network [35] computes dense flow correspondences. We further refine the relative pose with dense flow over the rigid region. With the refined pose, we compute 3D motion field and projected scene flow from the egomotion flow.

## 2   Related Work

**Scene Flow:** Scene flow estimation in dynamic scenes brings together fundamental computer vision algorithms in optical flow, and pose estimation of camera and objects. Vedula et al. [39] defined the 3D motion field as *scene flow*, and proposed a method to compute *dense non-rigid* 3D motion fields from a fixed multi-view set-up. Its extension to a moving camera case needs to disambiguate the camera ego-motion from object scene motions in 3D. Due to the intrinsic complexity of such task, existing methods often address it with known camera parameters [1,37] or assume scene motions are piecewise rigid [21,23,10,41,42,44]. When depth is known, scene flow can be more accurately estimated. Quiroga et al. estimates RGB-D scene flow as a rigid flow composited with a non-rigid 6DoF transforms [27]. Sun et al. estimates scene flow as a composition of finite rigid moving objects [34]. Jaimez et al. separately solve rigid region as visual odometry and non-rigid regions as moving clustered patches conditioned on rigidity segmentation [18]. They solve the rigidity segmentation based on the robust residuals of two frame alignment, similar to [25,20] for camera tracking in dynamic environments. All of these approaches use rigidity as a prior, but can fail as the complexity of the dynamic scene increases. None of these methods use learned models. We show that the 3D motion field can be more accurately estimated using learned models for rigidity and optical flow.

**Learning Camera Transform and Rigidity:** Recently, various learning-based methods have been introduced for the joint estimation of camera transform and depth (or rigid structure) [36,40,51], and rigid motion tracking [3]. Most of them assume that the scene is either static [36], quasi-static (scene motions are minimal and can be dealt as outliers) [51], or that the camera remains static when a rigid scene motion occurs [3]. More recently, a few approaches[47,49] demonstrated the importance of learning *rigidity* to handle dynamic scenes. Wulff et al.

[47] assume the rigidity can be learned by finetuning the semantic segmentation network from a single image, while we posit that rigidity correlates spatially to the epipolar geometry. Yin and Shi [49] unsupervised learn the non-rigid flow residual in the 3D urban scene. We are interested in more general dynamic scenes with unconstrained scene motions observed from moving cameras, and we address this by directly learning the per-pixel rigidity in the supervised manner which can generalize to unseen scenes.

## 3   Rigidity, Scene Flow and Moving Camera

We focus on solving for the 3D motion field in the physical scene observed from a moving camera, commonly termed as scene flow [18,39]. Here we define the relationship between 2D image correspondences and scene flow in physical 3D scenes with object motions and camera motion derived from relative camera poses between two temporal views.

Let $\mathbf{x}_t \in \mathbb{R}^3$ be the location of a point $\mathbf{x}$ on a non-rigid surface $\Omega_t$ of a moving object with respect to a fixed world coordinate system at time $t$. We define $\delta \mathbf{x}_{t \to t+1}$ as the 3D motion vector of $\mathbf{x}$ from time $t$ to time $t+1$, also referred as scene flow in this paper. When $\mathbf{x}_t$ is observed by a camera with known intrinsics, we define $\pi(\mathbf{x}_t)$ to be the projection of $\mathbf{x}_t$ to image coordinates $\mathbf{u}_t$, and $\pi^{-1}(\mathbf{u}_t, z_t)$ the inverse projection into 3D camera coordinates given the known depth $z_t$ in the camera reference frame.

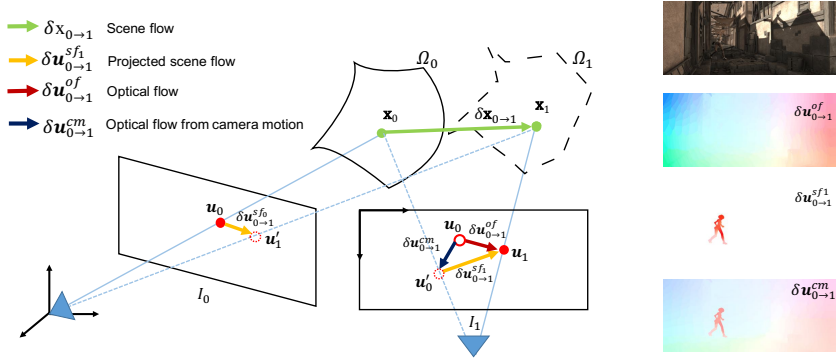

Fig. 3: **The geometry of two-frame scene flow**, where the camera moves from $I_0$ to $I_1$, and point $\mathbf{x}_0$ moves to $\mathbf{x}_1$ (green circles), and their projections in the two images are shown as $\mathbf{u}_0, \mathbf{u}_1$ respectively (red circles). Note that $\mathbf{u}'_0$ is a projected location of $\mathbf{x_0}$ in $I_1$, as if $\mathbf{x_0}$ were observed by $I_1$, and can be computed by camera motion as $\delta \mathbf{u}^{cm}_{0 \to 1}$, and $\mathbf{u}_0$ in $I_1$ is visualizing the pixel location it had in $I_0$. If the camera was static and observed both $\mathbf{x_0}$ and $\mathbf{x_1}$ at the position of $I_1$, optical flow $\delta \mathbf{u}^{of}_{0 \to 1}$ would be same to a projected scene flow $\delta \mathbf{u}^{sf_1}_{0 \to 1}$. The right image shows each flow in $I_1$ of dynamic scene under camera panning.

**Scene flow, 2D Optical Flow, and Camera Pose** Optical flow offers direct 2D associations of measurements in $I_t$ and $I_{t+1}$. Suppose $\mathcal{C}_t$ is the known camera extrinsics matrix for $I_t$, then the optical flow $\delta \mathbf{u}_{t \to t+1}$ from $I_t$ to $I_{t+1}$ can be defined as follows:

$$\delta \mathbf{u}^{of}_{t \to t+1} = \pi(\mathcal{C}_{t+1}(\mathbf{x}_t + \delta \mathbf{x}_{t \to t+1})) - \pi(\mathcal{C}_t \mathbf{x}_t) \qquad (1)$$

Equation 1 states the two-view geometric relationship between 2D optical flow and 3D scene flow. We can simplify it by considering the camera's relative motion from $I_0$ to $I_1$, i.e. assuming $t = 0$ and setting $\mathcal{C}_0$ to identity:

$$\delta \mathbf{u}^{of}_{0 \to 1} = \pi(\mathcal{C}_1(\mathbf{x}_0 + \delta \mathbf{x}_{0 \to 1})) - \pi(\mathbf{x}_0) \qquad (2)$$

Given the optical flow $\delta \mathbf{u}^{of}_{0 \to 1}$ and the depth from the RGBD data, the 3D scene flow vector can be computed as:

$$\delta \mathbf{x}_{0 \to 1} = \mathcal{C}_1^{-1} \pi^{-1}(\mathbf{u}_0 + \delta \mathbf{u}^{of}_{0 \to 1}, z_1) - \pi^{-1}(\mathbf{u}_0, z_0) \qquad (3)$$

Note that $\mathcal{C}_1$ can be computed from 2D correspondences that follow two-view epipolar geometry [13], and the corresponding points should lie on the rigid and static background structure. This is especially challenging when the scene contains dynamic components (moving objects) as well as a rigid and stationary background structure. As such, identifying inliers and outliers using *rigidity* is a key element for successful relative camera pose estimation, and thus is necessary to achieve reaching accurate scene flow estimation in a dynamic scene [18], which we will discuss in Sec. 4.

**Egomotion Flow from a Moving Camera in a Static Scene:** When an observed $\mathbf{x}$ in a scene remains static between the two frames, $\delta \mathbf{x}_{0 \to 1} = \mathbf{0}$ and therefore $\mathbf{x}_1 = \mathbf{x}_0$. Then, the observed optical flow is purely induced by the camera motion and we refer it as a camera egomotion flow:

$$\delta \mathbf{u}^{cm}_{0 \to 1} = \pi(\mathcal{C}_1 \mathbf{x}_0) - \pi(\mathbf{x}_0) \qquad (4)$$

**Projected Scene Flow and Rigidity:** As described in Fig. 3, the projected scene flow is a projection of a 3D scene flow $\delta \mathbf{x}_{0 \to 1}$ in $I_1$ if $\mathbf{x}_0$ was observed from $I_1$, which can be computed from camera ego-motion and optical flow:

$$\delta \mathbf{u}^{sf}_{0 \to 1} = \delta \mathbf{u}^{of}_{0 \to 1} - \delta \mathbf{u}^{cm}_{0 \to 1} \qquad (5)$$

The projected scene flow (in a novel view) is also referred as non-rigid residual [27,49]. All locations with zero values in projected scene flow indicate the rigidity region in ground truth data. As demonstrated in Fig. 3, the projected scene flow is a useful tool to evaluate the results of dense scene flow estimation in the 2D domain which requires accurate estimation of both camera pose and optical flow. Thus, we use it as the evaluation metric in Sec. 6.

## 4   3D Motion Field Estimation Pipeline

We introduce a framework that refines the relative camera transform and the optical flow with a rigidity mask for accurate scene flow estimation. Fig. 2 shows the overview of our proposed pipeline. Given a temporal pair of RGB-D images, we concurrently run the optical flow and rigidity-transform network. The optical flow network [35] offers the 2D correspondence association between frames, and our proposed rigidity-transform network provides an estimate of the camera transform and the rigidity mask.

### 4.1   Rigidity-Transform Network

Previous work on camera pose estimation using CNNs focused on either static or quasi-static scenes, where scene motions are absent or their amount is minimal [36,40,51]. In dynamic scenes with a moving camera, camera pose estimation can be challenging due to the ambiguity induced by the camera motion and scene (object) motion. Although existing approaches leverage prior information in motion or semantic knowledge [18,34,27,30] to disambiguate the two, the priors are usually not general for different scenes.

We propose to infer the rigidity from epipolar geometry by a fully-convolutional network that jointly learns camera motion and segmentation of the scene into dynamic and static regions from RGB-D inputs. We represent this rigidity segmentation as a binary mask with the static scene masked as rigid. The rigid scene components will obey the rigid transform constraints induced by the camera ego-motion and serve as the regions of *attention of the camera transform*. We name it rigidity-transform network (RTN), shown in Fig. 4.

**RTN:** Given a pair of RGB-D frames, we pre-process each frame into a 6 channel tensor $[(u-c_x)/f_x, (v-c_y)/f_y, 1/d, r, g, b]$, from camera intrinsic parameters $[f_x, f_y, c_x, c_y]$ and the depth $d$. Due to the range of depth values, this representation is numerical stable in training and delivers good generalization performance. We truncate $1/d$ to the range $[1e-4, 10]$, which can cover scenes of various scales. We concatenate the two-frame tensors to a 12-channel tensor as input to our network. The network is composed of an encoder followed by pose regression and a decoder followed by the rigidity segmentation.

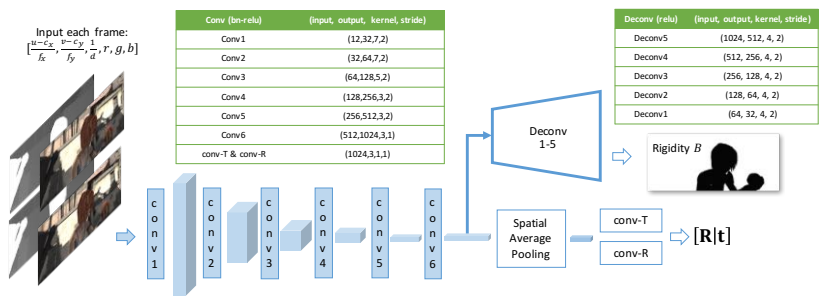

Fig. 4: **Rigidity-Transform network (RTN) architecture** The inputs to the RTN are 12 channel tensors encoded with $[(u - c_x)/f_x, (v - c_y)/f_y, 1/d, r, g, b]$ computed from a pair of RGB-D images and their intrinsics. It is a fully convolutional encoder-decoder architecture predicting pose as a translation and euler angles, and scene rigidity as a binary mask.

**Encoder:** We use five stride-2 conv-layers which gradually reduce spatial resolution and one stride-1 convolution as the conv-6 layer. Each convolution is followed by a batchnorm and ReLU layer. In the bottleneck layer, the target is to predict the camera relative translation $\mathbf{t}$ and rotation $\Theta$. After the conv-6 layer, we use a spatial-average pooling (SAP) to reduce the feature into a $1024D$ vector. With two $1 \times 1$ convolution layers that output 3 channels, we separately estimate the $\mathbf{t}$ and $\Theta$. We assume the relative camera transformation between two frames is small and thus we represent the rotation $\mathbf{R}(\alpha, \beta, \gamma) = \mathbf{R}_x(\alpha)\mathbf{R}_y(\beta)\mathbf{R}_z(\gamma)$ with Euler angles $\Theta = [\alpha, \beta, \gamma]$. The regression loss is a weighted combination of the robust Huber loss $\rho(\cdot)$ for translation and rotation as:

$$\mathcal{L}_p = \rho(\mathbf{t} - \mathbf{t}^\star) + w_\Theta \rho(\Theta - \Theta^\star) \qquad (6)$$

**Decoder:** The decoder network is composed of five deconvolution (transpose convolution) layers which gradually upsample the conv-6 feature into input image scale and reshape it into the original image resolution. We estimate the rigidity attention as a binary segmentation problem with binary cross-entropy loss $\mathcal{L}_r$. The overall loss is a weighted sum of both loss functions: $\mathcal{L}_c = w_p \mathcal{L}_p + \mathcal{L}_r$.

**Enforcing Learning from Two Views:** We enforce the network to capture both scene structures and epipolar constraints using two views rather than a single-view short-cut. First, our network is fully convolutional, and we regress the camera pose from the SAP layer which preserves feature distributions spatially. Features for rigidity segmentation and pose regression can interact directly with each other spatially across each feature map. We do not use any skip layer connections. Our experiments in Sec. 6 show that joint learning of camera pose and rigidity can help RTN to achieve better generalization in complex scenes. Second, we randomly use two identical views as input, and a fully rigid mask as output with 20% probability during data augmentation, which prevents the network from only using a single view for its prediction.

### 4.2 Pose Refinement from Rigidity and Flow

To solve for the 3D motion field accurately from two views from equation 3, we require a precise camera transformation. Moreover, the pose output from RTN may not always precisely generalize to new test scenes. To overcome this, we propose a refinement step based on the estimated rigidity $B$ and bidirectional dense optical flow $\delta \mathbf{u}_{0 \to 1}^{of}$ and $\delta \mathbf{u}_{1 \to 0}^{of}$ (with forward and backward pass). We view the estimation of $\mathcal{C}_1$ as a robust least square problem as:

$$\operatorname*{argmin}_{\mathcal{C}_1} \sum_{\{\mathbf{x}_0, \mathbf{x}_1\} \in \Omega(B)} [\mathbf{I}]\rho(\mathcal{C}_1 \mathbf{x}_0 - \mathbf{x}_1) \qquad (7)$$

where $\mathbf{x}_i = \pi^{-1}(\mathbf{u}_i, z_i)$ in all background regions $B$, predicted by the RTN. $[\mathbf{I}]$ is an Iverson bracket for all the inlier correspondences. We filter the inlier correspondences in several steps. We first use forward-backward consistency check for bidirectional optical flow with a threshold of 0.75 to remove all flow correspondences which are not consistent. The removed region approximates the occlusion map $O$. We use a morphological operator with patch size 10 to dilate $B$ and $O$ to further remove the outliers on boundaries. From all correspondences, we uniformly sample bidirectional flow correspondences with a stride of 4 and select 1e4 points among them that are closest to the camera viewpoint. These help to solves the optimization more efficiently and numerically stable. We also use the Huber norm $\rho(\cdot)$ as a robust way to handle the remaining outliers. We solve equation 7 efficiently via Gauss-Newton with $\mathcal{C}_1$ initialized from the RTN output. Note that in most cases correspondences are mostly accurate, the initialization step trivially helps but can also be replaced by an identity initialization.

## 5    REFRESH Dataset

Training our network requires a sufficient amount of dynamic RGB-D images over scenes and ground truth in the form of known camera pose, rigidity mask, and optical flow. However, acquiring such ground truth from the real-world data is difficult or even infeasible. Existing dataset acquisition tools include rendered animations like SINTEL[2] and Monka[22], and frames captured from games [28]. SINTEL [2] has a small number of frames, so we use it for testing instead of training. Most approaches render scenes using rigid 3D object models [33,8,22] with the concept. Among all existing tools and datasets, only Things3D[22] provides sufficient 3D training samples for learning 3D flow with moving camera ground truth. However, it only uses a small set of 3D objects with textured images at infinity as static scene context and rigid objects as the dynamic scene, which does not provide realistic 3D scene structure for motion estimation that can generalize well.

To overcome the dataset issue, we propose a semi-synthetic scene flow dataset: REal 3D from REconstruction with Synthetic Humans, which we name as **REFRESH**. For this task we leverage the success of state of the art 3D reconstruction systems [6,12,48], which directly provide dense 3D meshes and optimized camera trajectories. We use a pre-captured RGB-D dataset and create dynamic 4D scenes by rendering non-rigid 3D moving objects with pre-defined trajectories. We overlay synthetic objects over the original footage to obtain a composite image with the ground truth as shown in Fig. 5.

**Real 3D Reconstructed Scenes:** We use the 3D meshes created with BundleFusion [6]. The authors released eight reference 3D meshes with the 25K input RGB-D images, camera intrinsic and extrinsic parameters.

**Synthetic humans:** We create non-rigid scene elements with the method introduced in SURREAL [38]. Each synthetic body is created from realistic articulated human body models [19] and pose actions are from the CMU MoCap database [17] with more than 20K sequences of 23 action categories. The human textures are composed of SMPL CAESAR scans and real clothing registered with 4Cap [26]. We create each synthetic human with random gender, body shape, cloth texture, action and their positions in the 3D scene which guarantees the diversity of dynamic scenes. We control the visibility of human models along the trajectory by putting the pelvis point of each human model in the free space w.r.t. the ego-centric viewpoint from a selected frame along the trajectory. The free space is sampled by the corresponding depth. For every 100 frames, we select $n$ frames ($n$ sample from $\sim \mathcal{N}(15, 5)$) and insert $n$ human models into the scene.

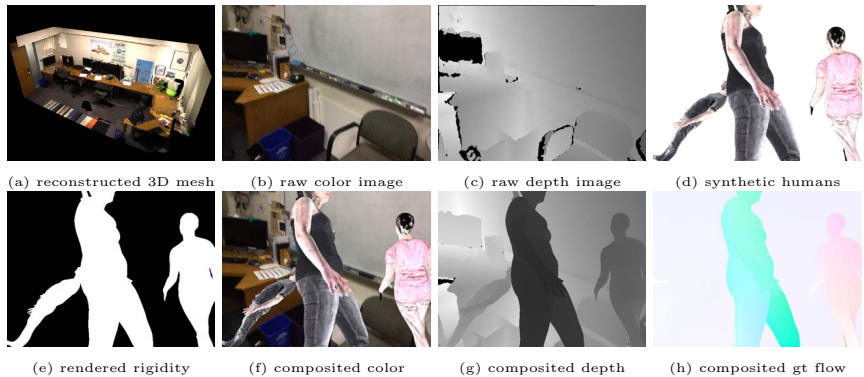

Fig. 5: **REFRESH dataset creation pipeline** With a captured RGB-D trajectory, the scene is reconstructed as a 3D mesh by BundleFusion [6] (a), with raw RGB-D input as (b) and (c). With sampled frames from the camera trajectory, we load synthetic human models [38] with motions randomly into the 3D as (d), and render the rigidity mask (e), Finally we composite the rendered synthetic ground truth with its corresponding rendered 3D views and the final semi-synthetic RGB-D views (f) and (h), with optical flow ground truth as (i).

**Rendering and ground-truth generation:** We use Cycles from the Blender [3] suite as our rendering engine. The lighting is created using spherical harmonics, as in Varol et al. [38]. First, we set the virtual camera using the same 3D scene camera intrinsic and spatial resolution. The camera extrinsic follows the real-data trajectory (computed from BundleFusion [6]). Thus, we can use the raw color image rather than rendered image as background texture which is photo-realistic and contains artifacts such as motion blur. With the same camera settings, we separately render the 3D reconstructed static mesh and the synthetic humans, and composite them using alpha-matting. Different from the color image, the depth map is rendered from the 3D mesh, which is less noisy and more complete than raw depth. Since the camera movement during the 3D acquisition is small between frames, we sub-sample frames at intervals of [1,2,5,10,20] to create larger motions. We employ a multi-pass rendering approach to generate depth, optical flow and rigidity mask as our ground truth.

---

[3] Blender: https://www.blender.org/

## 6   Experiments

We implemented the RTN in PyTorch, and the pose refinement in C++ with GTSAM 4.0 [7]. The PWCNet [35] is trained in Caffe. We integrate all the modules through Python. We use 68K images from our REFRESH dataset for training [4]. We train RTN from scratch using weight initialization from He et al.[14] and Adam optimizer ($\beta_1 = 0.9$ and $\beta_2 = 0.999$, learning rate of $2e^{-4}$) on 3 GPUs for 12 epochs. During training, the rigidity mask loss is accumulated over 5 different scales with balanced weights, and we choose $w_\Theta = 100$. We follow the same training as PWC-net Sun et al. [35]. We will release our code, datasets and REFRESH toolkit [5].

We evaluate our approach under various settings to show the performance of rigidity and pose estimation and their influence on scene flow estimation. For the effective analysis in scenes with different levels of non-rigid motions, we create a new test split from SINTEL data [2] based on the non-rigid number of pixels percentage. In Sec. 6.1, we provide a comparison of the performance with different settings for RTN, refinement and other state-of-the-arts methods. In Sec. 6.2, we qualitative evaluate of our method using real world images. Please also refer to our video for more qualitative evaluations.

### 6.1   Quantitative Evaluations

We first evaluate our approach on the challenging SINTEL dataset [2], which is a 3D rendered animation containing a sequence of 23 dynamic scenes with cinematic camera motion. The dataset has two versions with different rendering settings: *clean* and *final*. The latter set contains motion blur and depth of field effects, which are not present in the *clean* set. Since the official SINTEL test dataset does not provide RGB-D 3D flow evaluation, we split the SINTEL training set into train, validation, and test split. For the test split, to effectively evaluate and analyze the impact of different levels of non-rigid motions in the estimation, we choose *alley_2*(1.8%), *temp_2*(5.8%), *market_5*(27.04%), *ambush_6*(38.96%), *cave_4*(47.10%), where (·) indicates the average non-rigid regions in each scene sequence. These examples also contain a sufficient amount of camera motion. We use the first 5 frames in the rest of the 18 scenes as a validation set, and the remaining images for training in our finetuning setting.

We show our quantitative evaluations using flow metric in Table 1, relative pose metric in Table 2, and the rigidity IOU in Table 3. We list the end-point-error (EPE) in the ego-motion flow (EF) and projected scene flow (PSF) as defined in Sec.3. Our proposed metrics overcomes the traditional difficulty of 3D motion flow evaluation. We compare our method to two state-of-art optimization-based RGB-D scene flow solutions: SRSF [27] and VO-SF [18] which estimate the camera pose as part of the solution to flow correspondence.

---

[4] More details about the dataset are included in the supplementary materials.
[5] Code repository: https://github.com/NVlabs/learningrigidity.git

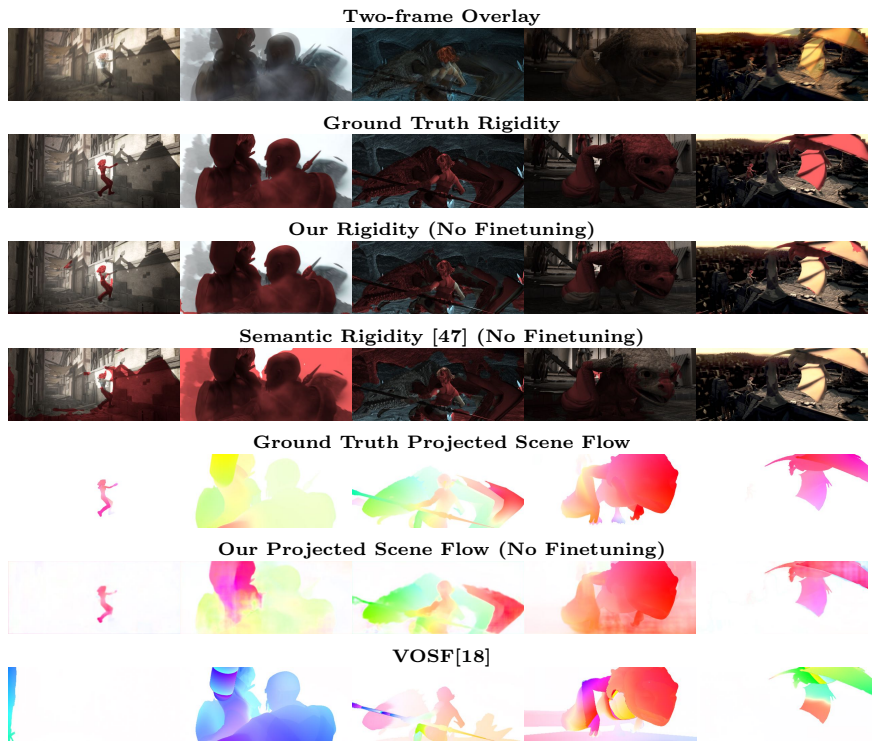

Fig. 6: **Qualitative visualization** on our SINTEL test split. We compare our rigidity prediction with the output using semantic rigidity [47] trained on our REFRESH dataset and our projected scene flow with output of VOSF [18].

In addition, we evaluate three types of baselines. The first one solves the refinement stage from flow without any inputs from RTN (*Refine Only*), which assumes rigidity often dominates the scene [25,18,20]. Secondly, we use three-point RANSAC to calculate the camera pose from the flow. Third, to fairly evaluate the rigidity of (RTN) and its generalization, we compare it to semantic rigidity estimation [47], which assumes that the non-rigid motion can be predicted from its semantic labeling. We follow Wulff et al [47] and use the DeepLab [5] architecture initialized from the pre-trained MS-COCO model, but trained over the same data we used for our model. In the pose refinement stage, we substitute our rigidity from RTN with the semantic rigidity. For the fine-tuned evaluation on SINTEL, we re-train both our RTN and the semantic rigidity network. All methods use the same optical flow network and weights, and all use the same depth from SINTEL ground truth. The qualitative comparison is shown in Fig.6.

**The Flow Metrics** in Table 1 show that: (1) compared to SRSF[27] and VOSF [18], our proposed algorithm with learned rigidity can improve scene flow accuracy by a significant margin with no further fine-tuning (NO FT) (rows (a),(b)vs(h); (k),(l)vs(r)); (2) the rigidity mask from our RTN performs bet-ter than the single-view semantic segmentation based approach [47], particularly in the more realistic *final pass* with no fine-tuning (row (d)vs(g),(h); (n)vs(q),(r)); (3) as shown in *RTN+refine*, the simultaneous learning of rigidity with pose transform achieves better generalization than learning rigidity alone (row (g)vs(h); (q)vs(r)); (4) RTN trained on our dataset generalizes better compared to the same RTN trained using Things3D[22] (row (f)vs(h); (p)vs(r)); and (5) the final approaches consistently outperforms all baselines. Note that the semantic rigidity [47] can achieve better performance (from Table 1) relying on fine-tuning on SINTEL, our learned rigidity can generalize to unseen complex scenes and perform as well as the fine-tuned model. Our rigidity prediction can capture unseen objects well, as shown by the dragon in Fig. 6.

Table 1: Quantitative evaluation in flow residuals using SINTEL dataset on our test split. The ratio of Nonrigid (NR) Region indicates the average ratio of pixels in the scene which represents the complexity of dynamic motion in the scene. We report the EPE in egomotion flow (EF) and projected scene flow (PSF). For all the baseline methods in both non-finetuning (NO FT) and finetuning (FT) setting, we use the same optical flow network trained as our method. The lowest residual under the same setting (e.g. NO FT, clean set) is highlighted as **bold**.

|  |  | NR Region<10% | | NR Region 10%-40% | | | | NR Region>40% | | All Test | |
|---|---|---|---|---|---|---|---|---|---|---|---|
|  |  | alley_2 | | temple_2 | | market_5 | | ambush_6 | | cave4 | | Average | |
|  |  | EF | PSF | EF | PSF | EF | PSF | EF | PSF | EF | PSF | EF | PSF |
|  | CLEAN (no motion blur) | | | | | | | | | | | | |
|  | (a) SRSF [27] | 4.24 | 7.25 | 7.59 | 16.55 | 25.26 | 31.67 | 17.84 | 37.21 | 10.77 | 11.82 | 12.47 | 18.57 |
|  | (b) VOSF [18] | 6.53 | 1.13 | 5.13 | 10.36 | 16.02 | 35.24 | 13.39 | 28.31 | 6.05 | 9.30 | 8.86 | 15.24 |
| NO FT | (c) Refine only | 0.29 | 0.48 | 0.90 | 2.95 | 8.81 | 22.34 | 3.59 | 14.39 | 2.18 | 5.88 | 3.09 | 8.47 |
|  | (d) Semantic[47]+Refine | 0.25 | 0.53 | 1.07 | 3.87 | 5.77 | 15.74 | 1.70 | 9.58 | 0.85 | 4.34 | 1.96 | 6.42 |
|  | (e) RANSAC+Flow | 0.31 | 0.57 | 0.47 | 2.73 | 7.36 | 19.19 | 3.86 | 14.89 | 2.17 | 5.94 | 2.69 | 7.78 |
|  | (f) RTN(use Things[22])+Refine | 0.34 | 0.60 | 1.47 | 3.98 | 7.21 | 18.73 | 21.84 | 23.97 | 1.17 | 4.90 | 4.20 | 5.85 |
|  | (g) RTN(no-pose)+Refine | **0.13** | **0.45** | 0.49 | 2.79 | 5.78 | 16.24 | 3.72 | 16.92 | 1.67 | 5.37 | 2.07 | 7.09 |
|  | (h) **RTN+Refine** | 0.18 | 0.48 | **0.46** | **2.72** | **1.61** | **11.86** | **0.97** | **8.61** | **0.63** | **4.05** | **0.74** | **5.10** |
| FT | (i) Semantic[47]+Refine | 0.19 | **0.46** | 0.50 | 2.73 | 2.73 | 13.45 | 1.13 | 9.94 | 2.07 | 5.87 | 1.35 | 5.98 |
|  | (j) **RTN+Refine** | **0.18** | 0.47 | **0.42** | **2.64** | 1.69 | 11.53 | 0.47 | 7.74 | 0.91 | 4.34 | 0.77 | 5.03 |
|  | FINAL (with motion blur) | | | | | | | | | | | | |
|  | (k) SRSF [27] | 4.33 | 7.78 | 7.59 | 15.51 | 24.93 | 31.29 | 17.26 | 39.08 | 10.80 | 13.29 | 12.37 | 18.86 |
|  | (l) VOSF [18] | 6.29 | 1.54 | 5.69 | 8.91 | 15.99 | 35.17 | 13.37 | 24.02 | 6.23 | 9.28 | 8.96 | 14.61 |
| NO FT | (m) Refine only | 0.28 | 0.57 | 0.90 | 3.77 | 8.80 | 20.64 | 3.59 | 20.41 | 2.18 | 6.52 | 3.09 | 8.95 |
|  | (n) Semantic[47]+refine | 0.25 | 0.52 | 0.96 | 3.83 | >100 | >100 | 20.23 | 35.46 | 11.05 | 12.81 | >100 | >100 |
|  | (o) RANSAC+Flow | 0.36 | 0.61 | 0.62 | 3.41 | 4.68 | 18.69 | 5.79 | 20.86 | 2.28 | 6.55 | 2.31 | 8.47 |
|  | (p) RTN(use Things[22])+Refine | 0.25 | 0.52 | 5.06 | 9.82 | 4.88 | 16.99 | 33.44 | 52.21 | 1.05 | 5.07 | 5.44 | 11.88 |
|  | (q) RTN(no-pose)+Refine | 0.19 | 0.48 | **0.82** | **3.58** | 2.15 | 13.97 | 3.34 | 20.02 | 1.52 | 5.72 | 1.36 | 7.14 |
|  | (r) **RTN+Refine** | **0.18** | **0.47** | 0.88 | 3.93 | **0.79** | **11.87** | **2.82** | **19.42** | **0.66** | **4.66** | **0.82** | **6.29** |
| FT | (s) Semantic[47]+Refine | **0.19** | **0.48** | 1.91 | 5.19 | 1.58 | 13.02 | 2.58 | 19.11 | 2.13 | 6.50 | 1.55 | 7.39 |
|  | (t) **RTN+Refine** | 0.21 | **0.48** | **0.66** | **3.27** | **0.97** | **11.35** | **2.34** | **19.08** | 0.74 | 4.75 | 0.79 | 6.12 |

**The Pose Metrics** evaluations in Table 2 include two other baselines: depth-based ORB-SLAM[24] and point cloud registration [50]. As mentioned, the ac-curacy of all relevant methods in dynamic scenes with moving camera highly relies on the ability ignore the non-rigid surfaces. As shown in the table, our pose directly predicted from RTN can achieve same or better accuracy with all relevant methods, and our final solution *without fine-tunning* can out-perform all state-of-the-art methods by a significant margin.

Table 2: Quantitative evaluation in relative camera transfrom using on our SINTEL test split. We report the relative pose error [32] (RPE) composed of translation (t) error and rotation error (r) in Euler angles (degree) in SINTEL depth metric averaged on from outputs using *clean* and *final* pass.

|  | NR Region <10% | | | | NR Region 10% - 40% | | | | NR Region >40% | | All Test | |
|---|---|---|---|---|---|---|---|---|---|---|---|---|
|  | alley_2 | | temple_2 | | market_5 | | ambush_6 | | cave4 | | AVERAGE | |
|  | RPE(t) | RPE(r) | RPE(t) | RPE(r) | RPE(t) | RPE(r) | RPE(t) | RPE(r) | RPE(t) | RPE(r) | RPE(t) | RPE(r) |
| ORB-SLAM [24] | 0.0300 | 0.0190 | 0.1740 | 0.0220 | 0.1500 | 0.0160 | 0.0550 | 0.0280 | 0.0167 | 0.0277 | 0.0894 | 0.0218 |
| SRSF [27] |  | 0.0487 | 0.0141 | 0.1763 | 0.0117 | 0.1566 | 0.0105 | 0.0672 | 0.0729 | 0.0218 | 0.0150 | 0.0980 | 0.0180 |
| VOSF[18] |  | 0.1043 | 0.0316 | 0.1055 | 0.0155 | 0.0605 | **0.0006** | 0.0375 | 0.0190 | 0.0438 | 0.0046 | 0.0750 | 0.0136 |
| Registration [1] |  | 0.0400 | 0.0094 | 0.3990 | 0.0381 | 0.0269 | 0.0073 | 0.0698 | 0.0225 | 0.0551 | 0.0076 | 0.1251 | 0.0162 |
| RANSAC+Flow |  | 0.0026 | 0.0047 | 0.0258 | 0.0033 | 0.0446 | 0.0043 | 0.0318 | 0.0082 | 0.0318 | 0.0411 | 0.0267 | 0.0039 |
| Our RTN Pose |  | 0.0349 | 0.0237 | 0.1589 | 0.0120 | 0.1520 | 0.0208 | 0.0455 | 0.0493 | 0.0233 | 0.0212 | 0.0883 | 0.0220 |
| Ours (no ft) |  | **0.0015** | **0.0036** | **0.0215** | **0.0010** | **0.0059** | 0.0009 | **0.0153** | **0.0061** | **0.0053** | **0.0009** | **0.0091** | **0.0020** |

Table 3: Evaluation of rigidity using mean IOU of rigid and nonrigid scenes.

| mean IOU | REFRESH val | SINTEL clean val | SINTEL final val |
|---|---|---|---|
| Semantic Rigidity [2] trained on REFRESH | 0.934 | 0.392 | 0.446 |
| RTN trained on Things [4] | - | 0.283 | 0.286 |
| RTN trained on our REFRESH | **0.956** | **0.542** | **0.627** |

**The Rigidity Metric** in Table 3 further shows the generalization in rigidity estimation. Our approach trained on our dataset generalizes significant better compared to the same approach trained using Things3D[22] and the semantic rigidity[47] using the same data.

### 6.2   Evaluation on Real-world Images

To test our algorithm in real-world dynamic scenarios, we use three TUM RGB-D sequences [32] which contains dynamic motions observed from a moving Kinect camera. The depth input is noisy with missing observations and the color images contain severe motion blur. We use the raw color and depth input with provided calibrated camera intrinsics as input, and mark the regions as invalid region when the depth value is not within $[0.1, 8]$. In invalid regions, we ignore the rigidity prediction and treat the flow correspondence as outliers. Considering there is no 3D motion flow ground truth for our real data, we visualize the rigidity prediction and projected scene flow to qualitatively show the performance in Fig. 7. Our results show that our trained model on semi-synthetic data can also generalize well to real noisy RGB-D data with significant motion blur.

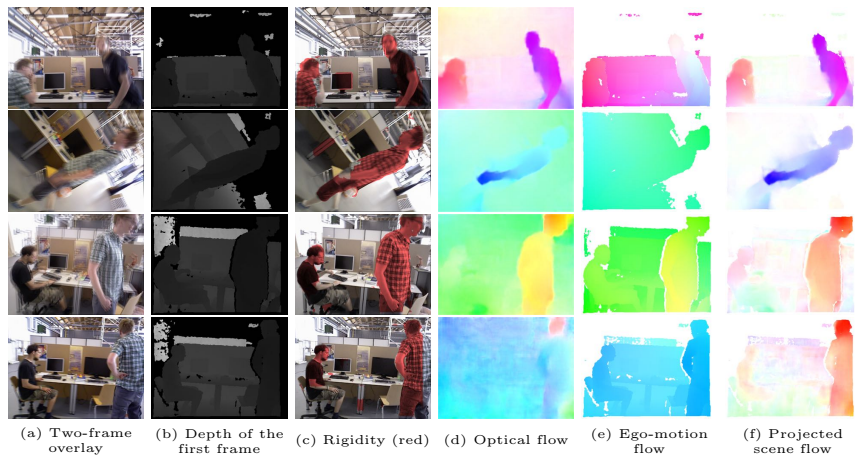

(a) Two-frame overlay   (b) Depth of the first frame   (c) Rigidity (red)   (d) Optical flow   (e) Ego-motion flow   (f) Projected scene flow

Fig. 7: **Qualitative visualization** of dynamic sequences in TUM [32] sequences.

## 7  Conclusion and Future Work

We have presented a learning-based approach to estimate the rigid regions in dynamic scenes observed by a moving camera. Furthermore, we have shown that our framework can accurately compute the 3D motion field (scene flow), and the relative camera transform between two views. To provide better supervision to the rigidity learning task and encourage the generalization of our model, we created a novel semi-synthetic dynamic scene dataset, REFRESH, which contains real-world background scenes together with synthetic foreground moving objects. Through various tests, we have shown that our proposed method can outperform state-of-the-art solutions. We also included a new guideline for dynamic scene evaluation regarding the amount of scene motion and camera motion.

We observed some cases where the rigidity mask deviates from the ground-truth. We noticed that in these situations the moving object size is small, or the temporal motions between the two frames are small. In these cases, the error and deviations scales are small, which does not significantly affect the 3D scene flow computed as a result. Note that the success of this method also depends on the accuracy of optical flow. In scenarios when the optical flow fails or produces a noisy result, the errors in the correspondences will also propagate to 3D motion field. In future work, we can address these problems by exploiting rendering more diverse datasets to encourage generalization in different scenes. We will also incorporate both rigidity and optical flow to refine the correspondence estimation and explore performance improvements with end-to-end learning, including correspondence refinement and depth estimation from RGB inputs.

**Acknowledgment** This work was partially supported by the National Science Foundation and National Robotics Initiative (Grant # IIS-1426998).

# Appendices

## A   Visualization for Qualitative Evaluation

**Color coding for flow vectors** We visualize the flow vectors in 2D following the color encoding in [2], for optical flow $\delta\mathbf{u}^{of}_{0\to1}$, egomotion flow $\delta\mathbf{u}^{cm}_{0\to1}$ and projected scene flow $\delta\mathbf{u}^{sf}_{0\to1}$. The central white color means there is no motion. Hue represents the flow vector direction, and color intensity represents the magnitude. All the flow vectors are normalized to the range [0,1] during visualization, shown in Figure 8. Thus, an accurate estimation of flow should have minimal difference w.r.t. ground truth flow visualization both in hue and intensity.

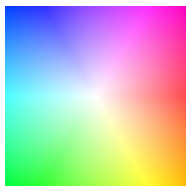

Fig. 8: **Flow color encoding** in all qualitative visualizations. The central white color means there is no motion. Hue represents the flow vector direction and color intensity represents the magnitude. All the flow vectors are normalized to the range of 0-1 during visualization.

We visualize the 3D dense scene flow following the same color encoding in 2D, simply by using the corresponding projected scene flow $\delta\mathbf{u}^{sf}_{0\to1}$ per-pixel. Such color encoding in projected image space can alleviate the noisy estimation for depth close to infinity, which usually has huge uncertainty in scale and thus affects the magnitude normalization.

## B   Training Optical Flow with REFRESH Dataset

We evaluate our optical flow model [35] trained on REFRESH dataset and compare it against models trained on FlyingChairs [8] and FlyingThings3D [22]. This evaluation serves as a sanity check of our dataset, and more importantly, an indication of its usefulness for scene flow.

Admittedly, the comparison with FlyingChairs [8] is not apple-to-apple. First, the FlyingChairs dataset is for 2D optical flow because it does not provide information such as depth, foreground masks, and camera ego-motion. More critically, the dataset has been tuned to match the statistics of the synthetic SINTEL dataset. However, it is important to check how valid our new dataset is for 2D optical flow, which is a sub-task of scene flow. As discussed earlier, the Fly-

Table 4: **Optical flow validation comparison** (EPE) on SINTEL[2] set (all images) using different datasets as training from scratch, validated at different number of training iterations. The same PWC-net [35] architecture is use in all training.

| SINTEL EPE (clean/final) | 6K | 30K | 60K | 90K | 120K |
|---|---|---|---|---|---|
| FlyingChairs[8] | 6.87/7.58 | 4.27/5.22 | 3.75/4.66 | 3.42/4.50 | 3.36/4.43 |
| FlyingThings3D [22] | 8.98/9.89 | 6.14/7.11 | 5.57/6.63 | 5.26/6.28 | 5.13/6.11 |
| **REFRESH (Ours)** | 5.82/6.54 | 4.27/5.23 | 3.85/4.78 | 3.45/4.48 | 3.42/4.46 |

ingThings3D dataset is the only training dataset that satisfies the requirements for training scene flow models[6].

As shown in Table 4, REFRESH dataset converges significantly faster and achieves better results on SINTEL than the FlyingThings3D dataset. The model trained on the REFRESH dataset also has similar performance as the one trained on the FlyingChairs dataset.

## C  Test Generalization to the Outdoor Domain

A fair quantitive evaluation on the KITTI dataset is challenging because: (1) the available ground truth depth from LIDAR is sparse for our method, and (2) the portion of moving regions is smaller. However, as an interest to see how our method and the data perform in a completely different domains with above domain discrepancies, we performed a qualitative evaluation on KITTI using the same RTN network trained on our dataset and dense depth calculated from PSMnet [4] output.

The rigidity results show that the RTN can generalize to KITTI reasonably well despite the domain gap and imperfect depth. We find the errors are more likely to happen in regions where the input depth uncertainty is higher and the surfaces are rigid planar, or textureless, which are not covered in our current generated data. This observation may inspire us to generate a mixture of nonrigid and rigid moving objects to improve the dataset diversity.

## D  REFRESH Datasets

### D.1  Dataset rendering details

The whole dataset creation is done using Blender 2.78[7], fully automated with python scripts without any GUI interaction, which scales well to the creation of the entire dataset. We separately render the background 3D meshes and foreground nonrigid humans, which allows us to speed up the rendering process. Since we use the raw color image as the background image and only use the geometry ground truth from multi-pass rendering (depth, flow, and segmentation), lighting does not affect background rendering with or without the foreground. Such separation can significantly boost the dataset creation speed. With a 28-core CPU server, we can finish the entire rendering process using BundleFusion [6] 3D scenes in two days.

---

[6] The Sintel dataset is held for validation and contains much fewer sequences to train scene flow models from scratch.
[7] Blender: https://www.blender.org/

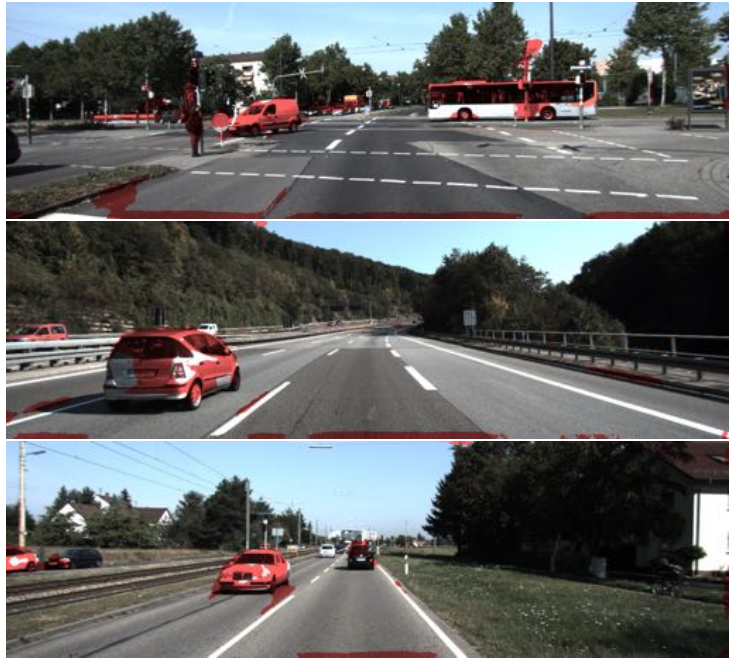

Fig. 9: Rigidity on KITTI with network trained on our REFRESH dataset. There is no finetuning on KITTI data.

**Background Static Mesh Rendering** Since we do not use the rendered color images in any process, we use a simplified setting for background rendering without ray-tracing, tile size as $512 \times 512$. The rendering time depends on the size of 3D mesh size. In average, we render one frame in ($< 1s$) in CPU, and we can finish the frame-by-frame rendering of 8 scenes of BundleFusion in 10 hours.

**Foreground Nonrigid Human Rendering** We create the human bodies following SURREAL [38] with synthetic textures (772 clothes textures and 158 CAESAR textures). The illuminated textures are used as the appearance of humans in our composted dynamic scenes. We use spherical harmonics with nine coefficients [11], with ambient illumination coefficient randomly sampled from [0.5, 1.5] and other coefficients randomly sampled from [-0.7, 0.7]. We implement this part by refactoring over [38], by extending SURREAL to arbitrary humans bodies with random textures and actions.

We split the camera trajectory into multiple clips. Each clip is a continuous 100-frame sequence, with randomly loaded human models and actions. There are two major motivations to rendering the outputs in clips rather than an entire trajectory: 1. We can load different random human bodies and motions for different clips in the same trajectory, which increase the motion diversity both in action and appearance; 2. There are numerous human models generated along the entire trajectory, which composes complex meshes in 3D and slow for rendering. Rendering individual clip with several human models is much faster in execution. We can render multiple pass image ground truth with an average of 3 seconds per frame.

**Create Ground Truth** We use Blender Cycles rendering passes to extract the per-pixel ground truth. We use the *Vector* node to retrieve the 2D vectors giving the frame by frame motions towards to the next and previous frame positions in pixel space, which are denoted as the forward/backward optical flow. Note that we currently do not retrieve the 3D motion vector representation of scene flow from Blender as one pass, which can be an extension to the current dataset in the future work.

We use the rendered depth from 3D scenes instead of the raw 3D scene depth for all the training. Compared to the raw depth, the rendered depth is less noisy and contains less missing measurements and has a per-pixel correspondence to the other ground truth, e.g., optical flow. However, the rendered depth does not guarantee a valid per-pixel value due to the incomplete 3D reconstruction from raw measurements. We marked the projected pixels from incomplete regions (holes in 3D reconstruction) as *invalid* region, and exclude them from the training on-the-fly.

### D.2   Dataset statistics

We rendered dataset using the optimized camera trajectory during 3D reconstruction as the camera extrinsic setting. Since the camera movement during 3D acquisition is small and stable between frames, we also use the sampled keyframes from the camera trajectory during rendering. We name the sub-sample trajectory based on their frame interval $n$ as *keyframe n*: $keyframe1$ represents that we use every frame along the trajectory during dataset creation and $keyframe10$ represents we use every ten frames. We list the number of static scene frames with varying keyframes in Table 5.

Table 5: The number of rendered images generated in our REFRESH dataset using BundleFusion [6] as 3D scenes.

|             | apt0  | apt1  | apt2 | copyroom | office0 | office1 | office2 | office3 | Total |
|-------------|-------|-------|------|----------|---------|---------|---------|---------|-------|
| keyframe 1  | 8560  | 8495  | 3873 | 4478     | 6083    | 5727    | 3494    | 3757    | 44467 |
| keyframe 2  | 4280  | 4248  | 1937 | 2239     | 3043    | 2863    | 1748    | 1882    | 22240 |
| keyframe 5  | 1712  | 1700  | 776  | 895      | 1220    | 1146    | 700     | 752     | 8901  |
| keyframe 10 | 856   | 849   | 338  | 447      | 609     | 572     | 349     | 376     | 4446  |
| keyframe 20 | 427   | 424   | 195  | 223      | 304     | 286     | 174     | 189     | 2222  |
| keyframe 50 | 171   | 169   | 78   | 89       | 123     | 114     | 69      | 75      | 888   |
| Total       | 16006 | 15885 | 7247 | 8371     | 11382   | 10708   | 6534    | 5149    | 83164 |

Table 6: Quantitative Evaluation on SINTEL dataset using all frames. All models in this evaluation are *not finetuned* and trained on REFRESH dataset. We report the EPE in egomotion flow (EF) and projected scene flow (PSF). The number in *failures* indicate the number of frames that has an EPE over 100, which is excluded in the EPE calculation. For all the baseline methods, we use the same optical flow network trained as our method. The lowest residual under the same setting (e.g. clean set) is highlighted as **bold**.

|  | Final Pass All | | | Clean Pass All | | |
| --- | --- | --- | --- | --- | --- | --- |
|  | EF | PSF | failures | EF | PSF | failures |
| Refine (from flow only) | 2.71 | 6.81 | 19 | 2.61 | 6.67 | 9 |
| Semantic rigidity [47] + refine | 6.19 | 9.35 | 25 | 4.57 | 7.68 | 12 |
| **RTN + Refine** | **1.78** | **5.81** | **17** | **1.75** | **5.72** | **6** |

Figure 10 shows the histogram distributions of our outputs in optical flow, depth, and rigidity from the rendered REFRESH dataset. We show the histogram distribution independently for the data rendered from different keyframes (1,2,5). Compare different keyframe splits, the distribution in depth and non-rigid area ratio in the images are similar and when using larger keyframes, the output optical flow tends to have a larger displacement. When using rendered outputs from larger keyframes, we can simulate the observations from a camera with larger motions.

During training, we empirically find the network generalize the best when using keyframe [1,2,5] from the optimized trajectory from BundleFusion. We use the first seven scenes in BundleFusion as our training set ('apt0', 'apt1', 'apt2', 'copyroom', 'office0', 'office1', 'office2') as our training set with a total of 69218 pairs of frames, and use 'office3' as the validation set with 6390 pairs of frames.

### D.3   Visualization

We visualize some examples of our datasets in Figure 11 across different scenes. The invalid regions are visualized as black in the depth image and white in the ground truth optical flow.

## E   Evaluation on SINTEL Dataset

### Quantitative Evaluation on Entire SINTEL Dataset

We evaluate our method using RTN and refine step on the all frames in entire SINTEL dataset compared to the two baseline methods in the paper as a supplement to the comparison in our test set split. First one is *refinement only*, which we denote as solving the refinement stage without any information acquired from RTN. Secondly, we compare our method to semantic rigidity estimation [47], which assumes that the non-rigid motion can be predicted from its semantic labeling. The semantic network is trained using the DeepLab [5] architecture with weights initialized from the pre-trained MS-COCO model on the same data we used for our model. In the pose refinement stage, we substitute our rigidity from RTN with the semantic rigidity. Both baselines use the same optical flow network with the same weights, and all methods use the same depth from SINTEL ground truth. We use the EPE in egomotion flow and projected scene flow as a metric. To exclude the effects of some catastrophic failures in some particular frames, we exclude those frames that have over 100 EPE values and separately count them as failure cases. None of the models are finetuned on SINTEL dataset. Table 6 shows the quantitative evaluation. It is worth to note that predicting rigidity based on semantics cannot generalize well across different domains, which can lead to bad rigidity localization that significantly harm the correspondence association. This evaluation also shows our method outperforms the two baseline methods.

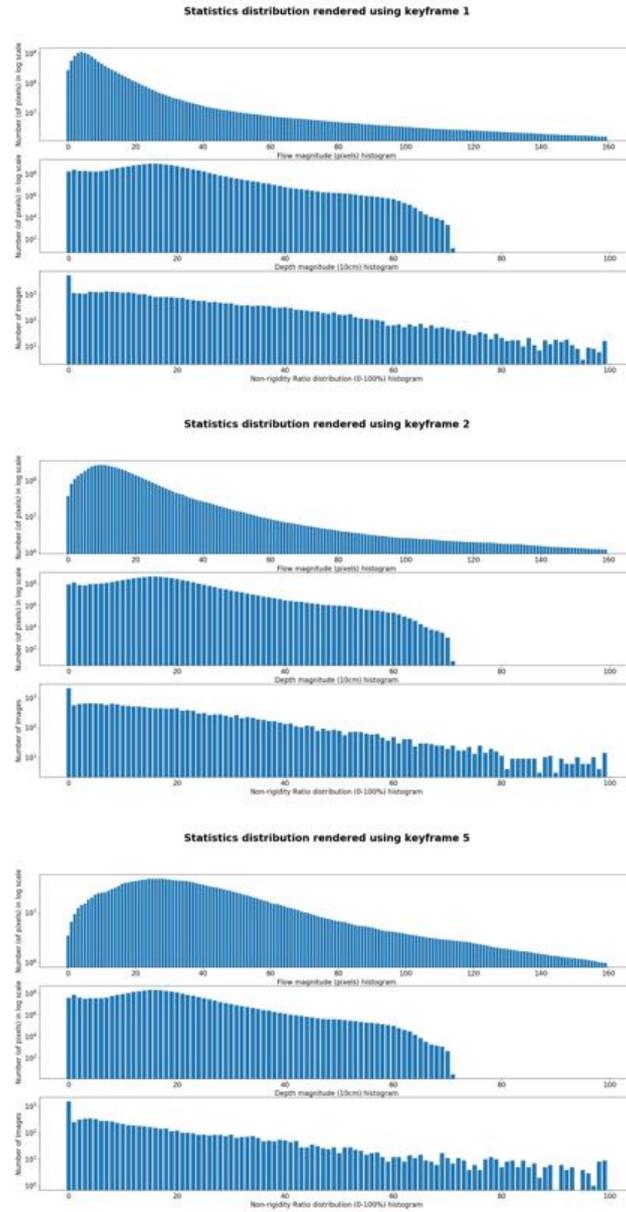

Fig. 10: Histogram distributions of optical flow, depth, and rigidity from our rendered REFRESH dataset in the training set. We calculate the distribution from three splits using keyframes 1, 2, 5 independently. In each of the split, we show the flow magnitude distribution (top) in pixels, depth distribution (medium) in centimeters, and nonrigid ratio (belows) in the number of different images.

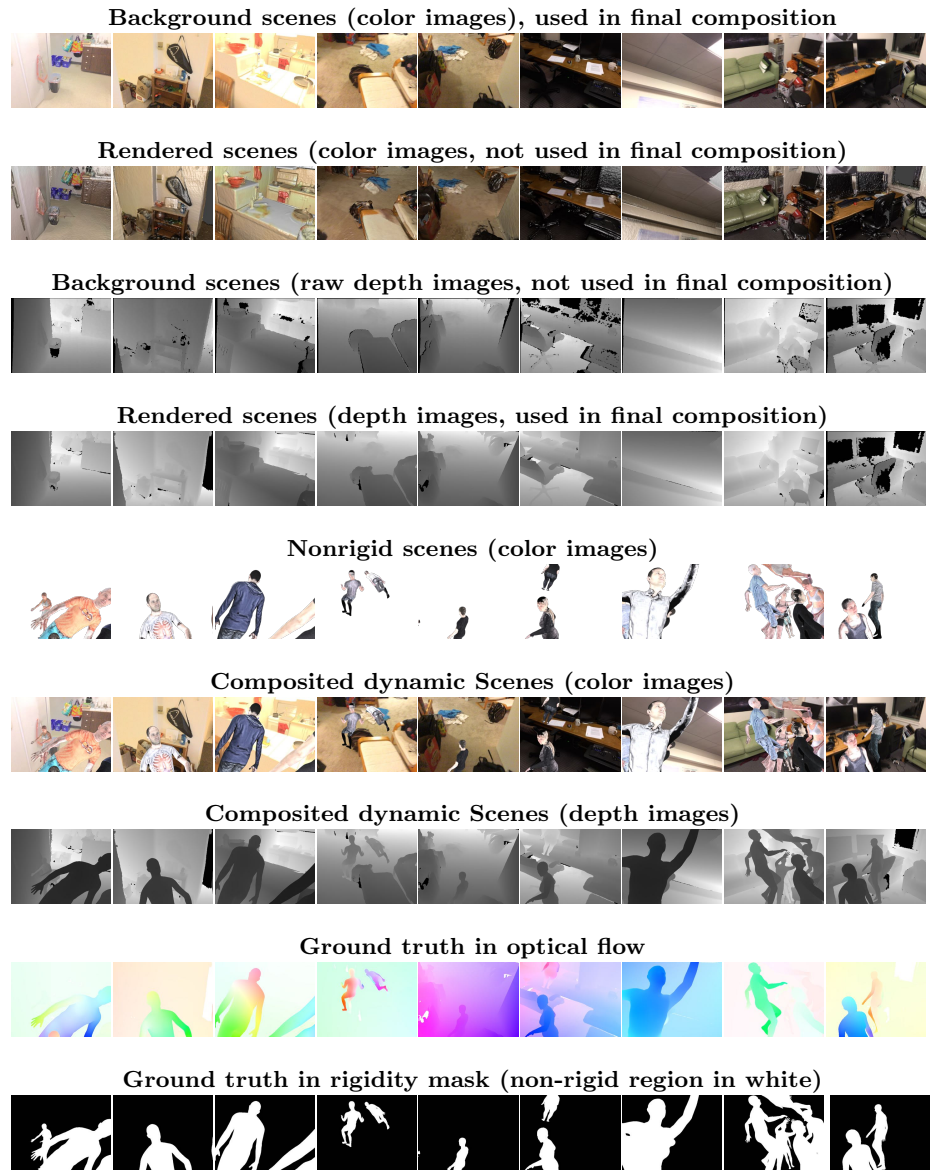

Fig. 11: **Qualitative visualization of Frames in REFRESH Datasets.**